\documentclass{llncs}
\usepackage{algorithm2e}
\usepackage{makeidx}  
\usepackage{tabularx}
\usepackage{graphicx}
\usepackage{amsmath}
\usepackage{amsfonts}

\usepackage[normalem]{ulem}
\usepackage{color}
\definecolor{ascol}{rgb}{0.7,0, 0}
\definecolor{mpcol}{rgb}{0,0, 0.7}

\definecolor{ascol}{rgb}{0.7,0, 0}

\begin{document}
\frontmatter          
\pagestyle{headings}  

\author{Evgenii Tsymbalov \and Maxim Panov \and
Alexander Shapeev}
\authorrunning{Evgenii Tsymbalov et al.} 
\institute{Skolkovo Institute of Science and Technology (Skoltech),\\ Nobel str. 3, Moscow, Russia, 121205, \\
\email{e.tsymbalov@skoltech.ru, m.panov@skoltech.ru, a.shapeev@skoltech.ru}
}
\title{Dropout-based Active Learning for Regression}
\maketitle              

\begin{abstract}
Active learning is relevant and challenging for high-dimensional regression models when the annotation of the samples is expensive. Yet most of the existing sampling methods cannot be applied to large-scale problems, consuming too much time for data processing. In this paper, we propose a fast active learning algorithm for regression, tailored for neural network models. It is based on uncertainty estimation from stochastic dropout output of the network. Experiments on both synthetic and real-world datasets show comparable or better performance (depending on the accuracy metric) as compared to the baselines. This approach can be generalized to other deep learning architectures.
It can be used to systematically improve a machine-learning model as it offers a computationally efficient way of sampling additional data.

\end{abstract}
\section{Introduction}

Active learning is crucial for the applications in which annotation of new data is expensive. 
Selection of samples is usually based on uncertainty estimation, by adding the samples on which the prediction is most uncertain to the training set.
The standard techniques include the models that estimate their uncertainty directly, such as Bayesian methods \cite{gp}, or algorithms that estimate uncertainty indirectly, through the deviation of the outputs of an ensemble of models. 
Models of the first type become computationally expensive when it comes to a large number of samples ($> 10^4$) and input dimensions ($> 10$). Ensemble-like models require full or partial independent training of several models, which is also time-consuming, even though it can be done in parallel. 

At the same time, neural networks can easily handle large amounts of data and thus are widely used in different areas of applied machine learning such as computer vision~\cite{szegedy2017inception}, speech recognition~\cite{sainath2013deep}, as well as physics~\cite{baldi}, manufacturing~\cite{anjos}, or chemistry~\cite{schutt}.
The popularity of deep learning has increased after the regularization techniques like dropout~\cite{hinton2012} and optimization procedures~\cite{rmsprop,adam} have been developed.
However, there is still a lack of theoretical understanding about capabilities of neural networks, in particular, estimating model uncertainty.

In our work, we propose a fast active learning algorithm for neural networks for regression problems. 
Our approach is based on a stochastic output of the neural network model, which is performed using different dropout masks at the prediction stage and used to rank unlabeled samples and to choose among them the ones with the highest uncertainty.
We demonstrate that our approach has a comparable or better performance with respect to the baselines in a series of numerical experiments.

The rest of the paper is organized as follows. We discuss the related work in Section~\ref{sect:related}. In Section~\ref{sect:methodology} our MCDUE (Monte-Carlo Dropout Uncertainty Estimation)-based active learning method is described in detail. Section~\ref{sect:results} contains numerical experiments on both real-world and synthetic data. Section~\ref{sect:summary} draws the conclusions to the paper.

\section{Related work}
\label{sect:related}

\subsection{Active learning}

\textit{Active learning}~\cite{settles} is a framework in which a machine-learning algorithm may choose the most informative unlabeled data samples, and ask an external oracle to annotate them. In statistics and engineering applications, such a setup is usually referred to as an \textit{adaptive design of experiments}~\cite{Fedorov1972,Forrester2008}.
In this setup, one has a set (finite or infinite) unlabeled examples, the so-called pool.
The function ranking the data points is referred to as an \textit{acquisition function} or a \textit{querying function}. Ideally, it should be designed in such a way that adding a relatively small number of data samples helps to improve the performance of the model.

There are numerous approaches to constructing an acquisition function. In the case of Bayesian models, such as the Gaussian Processes~\cite{gp}, model uncertainty may be defined explicitly: various estimates can be adopted to rank data points or even to choose them optimally over the defined region~\cite{Sacks1989,Burnaev2015}. However, the major drawback of such models is that calculations are usually intractable for a large number of input dimensions.
To cope with this problem various techniques and analogues were proposed, such as Bayesian Neural Networks~\cite{bnn}, but all of the methods still remain expensive to train when it comes to a considerably large train set or a complex model.

Another approach is a committee-based active learning~\cite{seung}, also known as \textit{query-by-committee}, where an ensemble of models is trained on the same or different parts of the data.
In this approach the measured inconsistency of the predictions obtained from different models for a given data sample may be considered as an overall ensemble uncertainty. 
Various investigations in this field include diversification of models~\cite{diverse_ensembles} and boosting/bagging exploitation~\cite{abe}. Unfortunately, in the case of neural network ensembles (see~\cite{li2018} for a detailed review) this approach often leads to an independent training of several models, which may be computationally expensive for the large-scale applications.

\subsection{Dropout}

Dropout~\cite{hinton2012,srivastava} is one of the most popular techniques used for a neural network regularization. To put it in plain words, it randomly mutes some of the neurons in hidden layers during the training stage, forcing them to output zero regardless of an input. This feature of the technique is taken into account during the backpropagation stage of training. Stochastic by nature and simple in implementation, it allows to efficiently reduce overfitting and thus has paved the way for new state-of-the-art results in almost every deep learning
application. First proposed as an engineering, empirical approach to reducing the correlation between weights, later it has obtained its theoretical interpretation as an averaged ensembling technique~\cite{srivastava}, a Bernoulli realization of the corresponding Bayesian neural network~\cite{gal2015} and a latent variable model~\cite{maeda2014}. It was shown in~\cite{gal_thesis} that using dropout at the prediction stage (i.e., stochastic forward passes of the test samples through the network, also referred to as \textit{MC dropout}) leads to unbiased Monte-Carlo estimates of the mean and the variance for the corresponding Bayesian neural network trained using variational inference. In \cite{gal_thesis}, Gal also proposes and analyses direct estimates for a model uncertainty. We use a lightweight version of this approach, estimating the model uncertainty based on a sample standard deviation (which is, up to a factor, an unbiased model variance estimate) of the stochastic output of the network.

Although there are some applications of the abovementioned approach to active learning setting for classification problems~\cite{cl1,cl2}, to the best of our knowledge, there has been no study of applicability of this approach to the regression task. In this work, we evaluate our MC dropout-based approach and compare it with high-throughput baselines. Our experiments have shown comparable or better accuracy and supremacy in a speed-accuracy trade-off of the proposed approach.

\section{Methodology}\label{sect:methodology}

\subsection{Problem statement}

Let
\begin{equation*}
    y = f(x),\  x \in \mathcal{X} \subset \mathbb{R}^n,\  y \in \mathbb{R}
\end{equation*} 
be some unknown function which is to be approximated
 using the values from the training set 
\begin{equation*}
    D_{\rm{train}} = \{ x^{\rm{train}}_j, f(x^{\rm{train}}_j) ,\  j = 1, \ldots, N_{\rm{train}}\}.
\end{equation*}
%
Suppose we have a model (more specifically, a neural network) $\hat{f}\colon \mathcal{X} \rightarrow \mathbb{R}$ trained on $D_{\rm{train}}$ with a mean squared error
\begin{equation*}
   L(\hat{f}, D_{\rm{train}}) = \sum_{j = 1}^{N_{\rm{train}}} \bigl(f(x^{\rm{train}}_j) - \hat{f}(x^{\rm{train}}_j)\bigr)^2
\end{equation*}
as a fitting criterion.

Let us now focus on the setting in which we want to decrease the value of loss function on some set of test points $D_{\rm{test}}$ by extending the training set $D_{\rm{train}}$ with some set of additional samples and performing an additional training of the model $\hat{f}$. 
More precisely, we are given another set of points called the ``pool''
\begin{equation*}
   \mathcal{P} = \{ x_j ,\  j = 1, \ldots, N_{\rm{pool}},\ \mathcal{P} \subset \mathcal{X}\},
\end{equation*}
which represents unlabeled data. Each point $x^* \in \mathcal{P}$ may be annotated by computing $f(x^*)$ so that the pair $\{x^*, f(x^*)\}$ is added to the training set. We suppose that in a practical application the process of annotation is expensive (i.e., it requires additional resources such as computational time or money), hence we need to choose as few additional points from the set $\mathcal{P}$ as possible to achieve the desired quality of the model. To achieve this goal, we use an acquisition function
\begin{equation*}
   A(\hat{f}, \mathcal{P}, D_{\rm{train}})\colon \mathcal{P} \rightarrow \mathbb{R}_+,
\end{equation*}
which ranks the points from $\mathcal{P}$ in such a way that the points with the larger values of $A$ become more appealing for the model to learn on. In the experimental setting of this work, we use a fixed number $m$ of the points to add at each stage of the active learning process. 

In most practical applications an acquisition function is related to a model uncertainty, which may be defined in various ways depending on the model and the field. There are also approaches, such as random sampling, that do not use the information from the model $\hat{f}$ (see Section~\ref{subsect:baselines} for further details). As for computationally heavy models with hundreds of thousands of parameters, such as neural networks, these approaches may be considered as more preferable. In the next section, we introduce a Monte-Carlo Dropout Uncertainty Estimation (MCDUE) approach, which enables us to collect uncertainty information from the neural network. 

\subsection{Monte-Carlo Dropout Uncertainty Estimation}

\begin{figure}[h!]
        \centering
        \includegraphics[scale=.5]{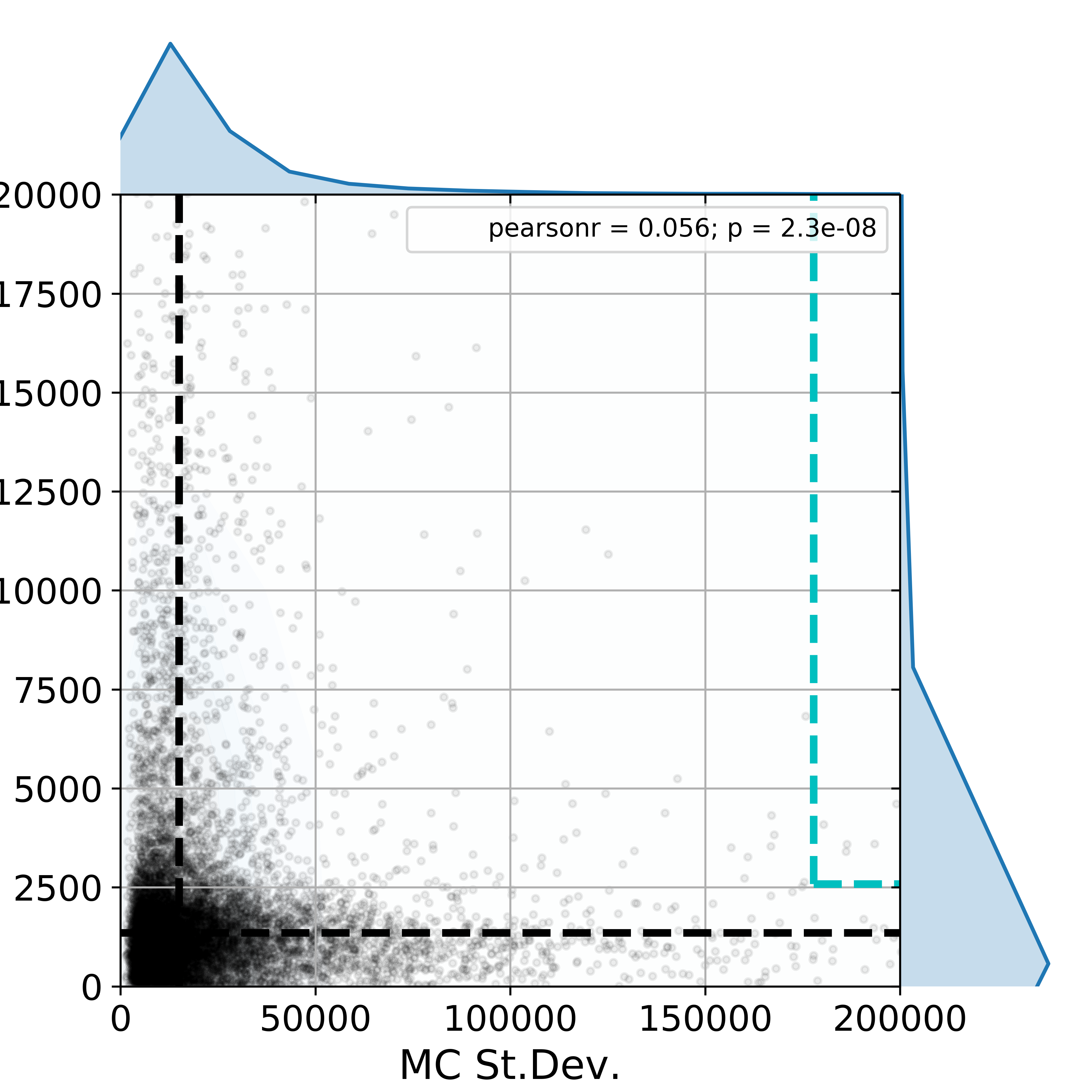} 
        \caption{Scatter plot shows the relation between the MC standard deviation and the absolute error for test samples. 
        The black dashed lines correspond to the medians of distributions, the vertical blue line corresponds to the 0.99 percentile of the MC standard deviation distribution, while the horizontal blue line shows the median percentile of the absolute error distribution of corresponding samples, which is equal to 0.783 in this case. 
        Five-layer neural network with a 256-128-64 structure was used on the Online News Popularity dataset \cite{fernandes2015}. 
        The Pearson correlation coefficient equals to 0.056, thus showing no linear relation between the absolute error and the MC standard deviation.}\label{fig:hist_worse}
\end{figure}

\begin{figure}[h!]
    \centering
        \centering
        \includegraphics[scale=.5]{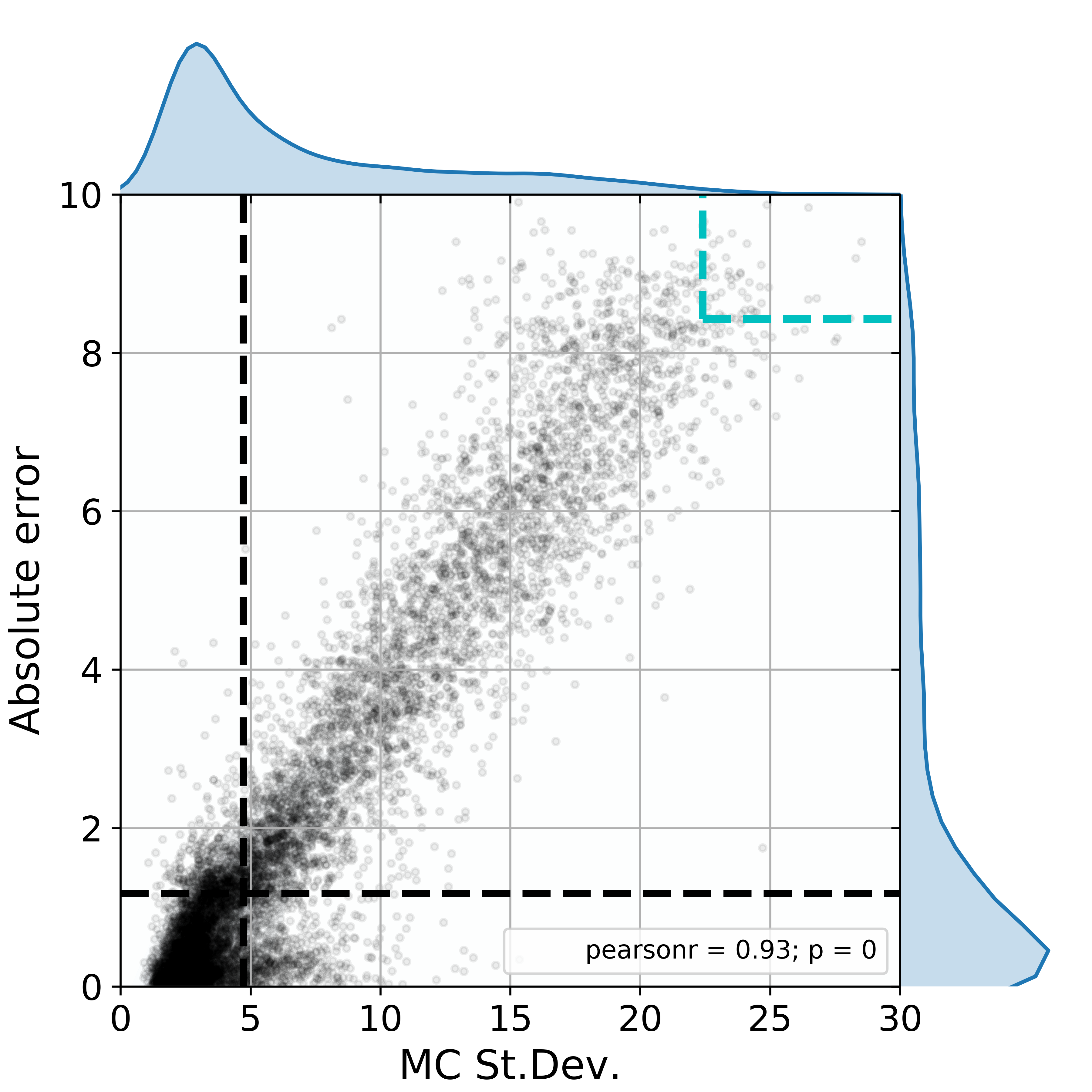} 
        \caption{
        Scatter plot shows the relation between the MC standard deviation and the absolute error for test samples. 
        The black dashed lines correspond to the medians of distributions, the vertical blue line corresponds to the 0.99 percentile of the MC standard deviation distribution, while the horizontal blue line shows the median percentile of the absolute error distribution of corresponding samples, which is equal to 0.975 in this case. 
        Five-layer neural network with a 256-128-64 structure was used on the CT slices dataset \cite{graf2011}. 
        The Pearson correlation coefficient equals to 0.93, thus showing an almost linear relation between the absolute error and the MC standard deviation, so if we choose a sample with a relatively high MC standard deviation, it will probably have large absolute error.}\label{fig:hist_good}
\end{figure}

Using dropout at the prediction stage allows us to generate stochastic predictions and, consequently, to estimate the variance of these predictions. Our approach rests on the hypothesis that data samples with higher standard deviations have larger errors of true function predictions. Although this is not always the case (see Figure~\ref{fig:hist_worse}), concerning a neural network of a reasonable size trained on a reasonable number of samples we have observed a clear correlation between dropout-based variance estimates and prediction errors (see Figure~\ref{fig:hist_good}). It should be noted that the result does vary (like any other result of neural network training) depending on several factors: architecture and size of the neural network, samples used for initial training and training hyperparameters, such as regularization, learning rate, and dropout probability. The MCDUE-based active learning algorithm we propose 
is summarized below.

\begin{enumerate}
    \item \textbf{Initialization}. Choose a trained neural network $\hat{f}(x) = \hat{f}(x, \omega)$, where \(\omega\) is a vector of weights. Set the dropout probability $\pi$. Set the number of stochastic runs $T$.
    \item \textbf{Variance estimation}. For each sample $x_j$ from the pool $\mathcal{P}$:
    \begin{enumerate}
        \item Make $T$ stochastic runs using dropout of the model $\hat{f}$ and collect outputs $y_k = \hat{f}_{k}(x_j) = \hat{f}(x_j, \omega_k), k = 1, \ldots, T$, where $\omega_k$ are sampled from Bernoulli distribution with parameter $\pi$.
        \item Calculate the standard deviation (as an acquisition function): 
        \begin{equation*}
        s_j = A^{\rm{MCDUE}}(x_j) = \sqrt{\frac{1}{T - 1} \sum_{k = 1}^T (y_k - \bar{y})^2},\ \bar{y} = \frac{1}{T} \sum_{k = 1}^T y_k.
    \end{equation*}
    \end{enumerate}
    \item \textbf{Sampling}. Pick $m$ samples with the largest standard deviations  $s_j$.
\end{enumerate}

In our experiments we used the dropout probability $\pi = 0.5$, and number of stochastic runs for each sample $T = 25$. We found out that, generally, the decrease in $\pi$ results in an approximately linear scaling of standard deviations $s_j$ for not too small $\pi$. The computational cost per one sample is $O(T N_{\rm{pool}})$. However, on modern GPU-based implementations the sampling can be done in parallel. One could also decrease $T$ to speed up the procedure. It should be emphasized that we can start with a pre-trained neural network from the previous iteration if we train the model on the extended training set; such a method may significantly speed-up retraining.


\subsection{Baselines}
\label{subsect:baselines}
We use the following baselines to compare the performance. \\
\textbf{Random sampling.} This algorithm samples random points $x$ from the pool. 
Computational cost is $O(1)$ in this case, which makes this algorithm the fastest compared to all the others.
\\
\textbf{Greedy max-min sampling.} To sample a point, this algorithm takes the point from the pool most distant from the training set (in the $l_2$ sense) and adds it to the training set. This process continues until the required number of points is added to the training set. The acquisition function for max-min sampling is
\begin{equation*}
    A_{\rm{MM}}(x) = \min_k ~ \|x - x^{\rm{train}}_k\|^2.
\end{equation*}
In this case, computational cost can be estimated as $O(N_{train} N_{pool})$.
\\
\textbf{Batch max-min sampling.} Although straightforward and intuitive, the max-min sampling is also computationally expensive in the case of a large number of dimensions and pool/training set size, since it requires that a full distance matrix is calculated on every stage of active learning. We propose the batch version that has the same acquisition function $A_{\rm{MM}}$ but samples $K$ points which are the most distant from the training set. Although it does seem less optimal, this solution speeds up sampling up to $K$ times assuming $K$ is the number of samples to be sampled on each iteration. In our experiments, we set $K$ equal to 4.

\section{Results}\label{sect:results}

\subsection{Experimental setup}

We focus on the non-Bayesian methods as comparable in terms of the computational time for large datasets and models and compare the active learning algorithms in the following experimental setup: 
\begin{enumerate}
    \item Initialization of the initial dataset $I$, training pool $\mathcal{P}$, number of samples added on each step $m$, the final size of the dataset $f$, network architecture, and learning parameters.
    \item The network is trained on the initial dataset $I$. Its weights are copied to the networks corresponding to each active learning algorithm.
    \item For each active learning algorithm:
    \begin{enumerate}
        \item While $|I| < f$:
            \begin{enumerate}
                \item Obtain the rank $r_j$ for every $x_j \in \mathcal{P}$ using an acquisition function $A$.
                \item Sample point set $S \subset \mathcal{P},\ |S| = m$ with maximal ranks $r_j$.
                \item Add $S$ to $I$: $I := I \bigcup S$.
                \item Exclude $S$ from the corresponding $\mathcal{P}$: $\mathcal{P} := \mathcal{P} / S$.
                \item Train the neural network on $I$.
            \end{enumerate} 
    \item Calculate the metrics.
    \end{enumerate}
            
\end{enumerate}

For each experiment, the number of training epochs was set to 10000. We used the $l_2$-regularization of the weights with the regularization parameter $\alpha = 10^{-5}$, a five-layer fully-connected network with the $256-128-64$ architecture and leaky linear rectifier~\cite{leaky_relu} with leakiness $\beta = 0.01$ as an activation function. We used the Theano library~\cite{theano} and Lasagne framework~\cite{lasagne}. Data points were shuffled and split in the following ratio: 20\% on a training set, 60\% on a pool, 20\% on the test set.

\subsection{Metrics}

During the neural network training, we optimize the mean squared error (MSE, $l_2$) metric. However, we also report the mean absolute error (MAE, $l_1$) and maximum absolute error (MaxAE, $l_\infty$). 

We believe that the actual task of regression is more general than optimization of one given metric (e.g., MSE); thus, the choice of a particular metric is merely an operationalization of the real problem behind the regression task.
Moreover, several applications exist, in which the maximal error is a much more appropriate accuracy metric (like chemistry or physical simulations) than the mean error. Unfortunately, it is hard to use the $l_\infty$ loss function for training neural network since it is non-differentiable. In case one deals with two algorithms that have a similar MSE and significantly different MaxAE, the algorithm with a smaller maximal error should be preferred. 

\subsection{Datasets}

We took the data from UCI ML repository \cite{uci}, see the Table \ref{tab:datasets} for more details. All the datasets represent real-world problems with $15+$ dimensions and $30000+$ samples. The exception is the synthetic Rosenbrock 2000D dataset, which has $10000$ samples and $2000$ dimensions.

\begin{table}[h]
\centering
\caption{Summary of the datasets used in our experiments.}
\label{tab:datasets}
\scalebox{0.77}{
\begin{tabular}{|c|c|c|c|c|c|c|}
\hline
\textbf{Dataset name} & \textbf{\# of samples} & \textbf{\# of attributes} & \textbf{Feature to predict} \\ \hline
BlogFeedback \cite{buza2014} & 60021 & 281 & Number of comments \\ \hline
SGEMM GPU \cite{nugteren2015} & 241600 & 18 & Median calculation time \\ \hline
YearPredictionMSD \cite{bm2011} & 515345 & 90 & Year \\ \hline
Relative location of CT slices \cite{graf2011} & 53500 & 386 & Relative location \\ \hline
Online News Popularity \cite{fernandes2015} & 39797 & 61 & Number of shares \\ \hline
KEGG Network \cite{shannon2003cytoscape} & 53414 & 24 & Clustering coefficient \\ \hline
Rosenbrock 2000D \cite{rosenbrock} & 10000 & 2000 & Function value \\ \hline
\end{tabular}
}
\end{table}

Since the neural network training procedure is stochastic by its nature, we conducted 20 experiments shuffling the dataset and re-initializing the weights each time. 

\subsection{Ratio plots}

\begin{figure}[h!]
    \hspace*{-1cm} 
    \includegraphics[scale=.45]{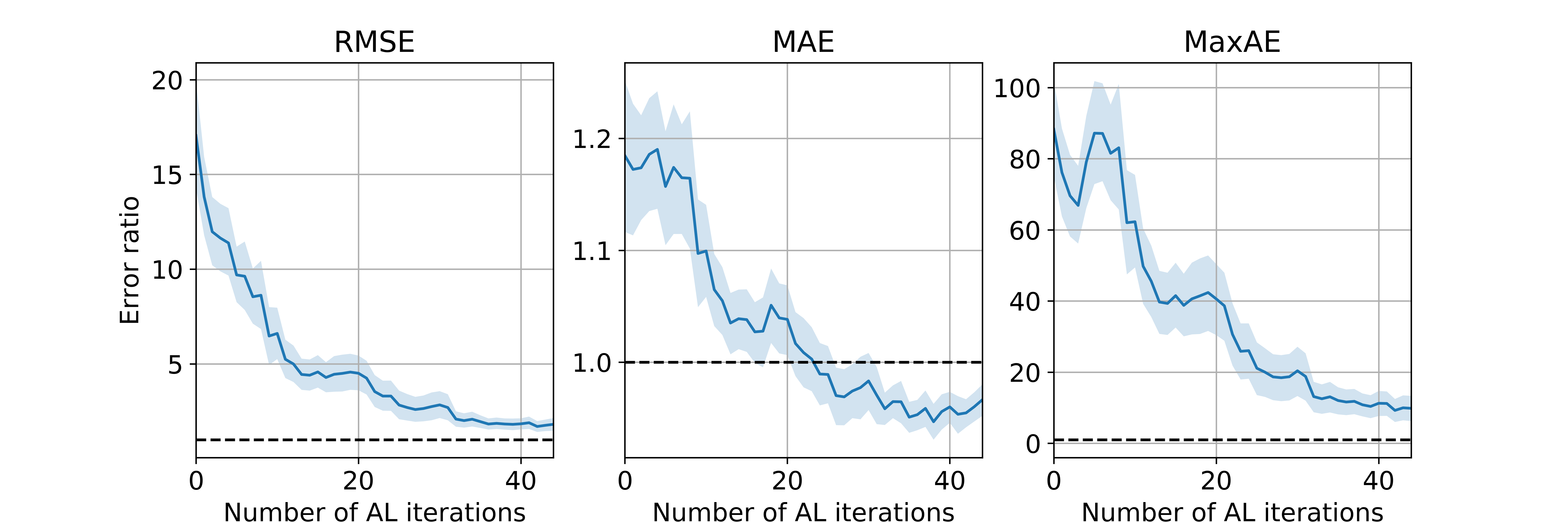}
        \caption{Comparison of MCDUE-based algorithm and random sampling algorithm: the ratio of errors (training curves) for various metrics on the KEGG Network dataset \cite{shannon2003cytoscape}.
        Ratio bigger than 1 (dashed line) shows the superiority of the MCDUE-based algorithm.
        The blue line shows the mean over 25 experiments, the standard deviation is also shown. 
        One can see that the proposed algorithm outperforms the random sampling significantly on RMSE and MaxAE metrics.
        }
    \label{fig:rnd_ratios}
\end{figure}

\begin{figure}[h!]
    \hspace*{-1cm} 
    \includegraphics[scale=.45]{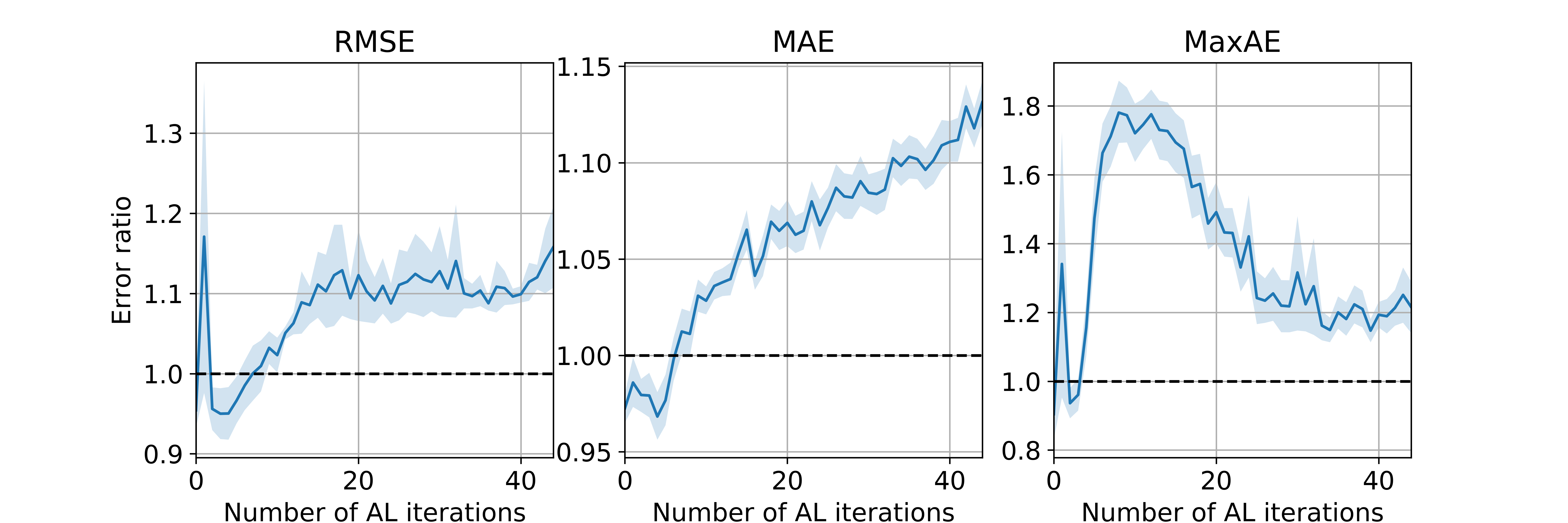} 
        \caption{Comparison of MCDUE-based algorithm and  max-min sampling algorithm: the ratio of errors (training curves) for various metrics on KEGG Network dataset \cite{shannon2003cytoscape}.
        Ratio bigger than 1 (dashed line) shows the superiority of the MCDUE-based algorithm.
        The blue line shows the mean over 25 experiments, the standard deviation is also shown. 
        One can see that the proposed algorithm slightly outperforms the max-min sampling across all the metrics by up to 20\%.}
    \label{fig:maxmin_ratios}
\end{figure}

First, we compare our MCDUE-based approach with the baseline approaches using the ratio of errors in various metrics.
Figures \ref{fig:rnd_ratios} and \ref{fig:maxmin_ratios} show that our active learning approach has a better performance than random sampling in RMSE and MaxAE metrics, and a small accuracy increase as compared to a max-min algorithm. It should be noted that as the number of active learning iterations (new data gathering and learning on the top of it) increases (thus leaving the data pool empty), ratio turns to 1.

\subsection{Dolan-More plots}

To compare the performance of the algorithms across the different datasets (see Table \ref{tab:datasets}), the initial number of training samples and training samples themselves, we will use Dolan-More curves, which, following \cite{dolan}, may be defined as follows. Let $q^p_a$ be an error measure of the $a$-th algorithm on the $\mathcal{P}$-th problem. Then, defining the performance ratio $r^p_a = \frac{q^p_a}{\min_x(q^p_x)}$, we can define the Dolan-More curve as a function of the performance ratio factor $\tau$:
\begin{equation}
    \rho_a(\tau) = \frac{\#(p: r^p_a \leq \tau)}{n_p},
\end{equation}
where $n_p$ is a total number of evaluations for the problem $p$. Thus, $\rho_a(\tau)$ defines the fraction of problems in which the $a$-th algorithm has the error not more than $\tau$ times bigger than the best competitor in the chosen performance metric.

\begin{figure}[h!]
    \hspace*{-1cm} 
    \centering
    \includegraphics[scale=.35]{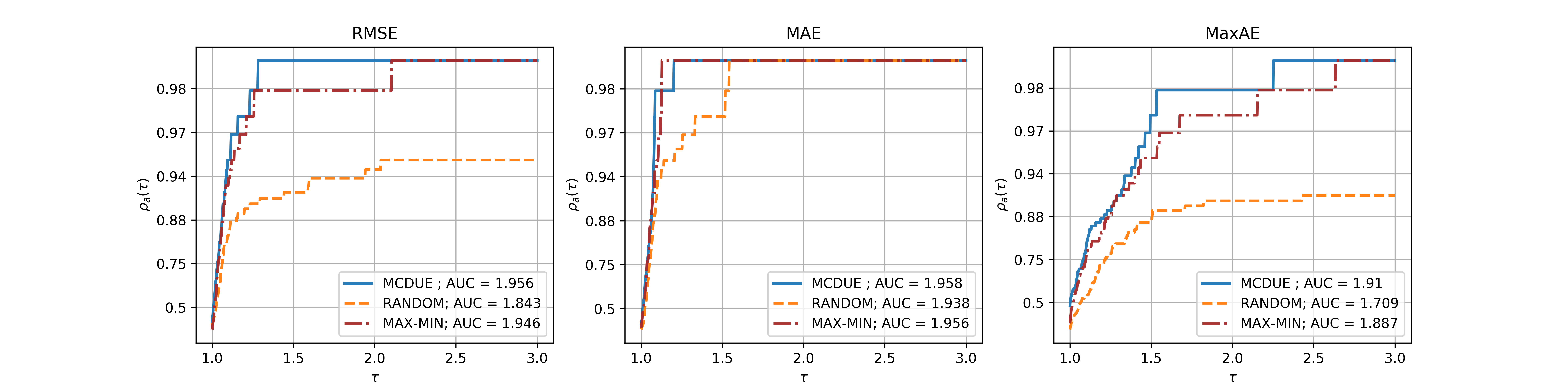} 
        \caption{Log-scaled Dolan-More curves for various metrics and acquisition functions. Figures on the legend indicate the area under the curve (AUC) metric for each algorithm. The number of training samples was chosen randomly from 1000 to the 20\% of the training set, the number $m$ of points to sample from the pool $\mathcal{P}$ was chosen randomly from 100 to 1100, with a 140 experiments conducted in total. The MCDUE-based approach outperforms the random sampling and is slightly better than batch max-min sampling.}
    \label{fig:dolan-more}
\end{figure}

We conducted a number of experiments with one iteration of active learning performed on the datasets from Table \ref{tab:datasets}, see Figure \ref{fig:dolan-more} for Dolan-More curves. Note that $\rho_a(1)$ is the ratio of problems on which the $a$-th algorithm performance was the best, and it is always the case of the MCDUE-based algorithm. Judging by the area under curve (AUC) metric, the MCDUE-based approach outperforms the random sampling and is slightly better than a batch max-min sampling.

\section{Summary and discussion}\label{sect:summary}

We have proposed an MCDUE-based (Monte-Carlo Dropout Uncertainty Estimation-based) approach to active learning for the regression problems. This approach allows neural network models to estimate self-uncertainty of unlabeled data samples.
Numerical experiments on real-life datasets have shown that our algorithm outperforms both random sampling and the max-min baseline.
Compared to the latter, our algorithm is faster and relies on the information provided by the model.

Theoretical connections with Bayesian neural networks require further investigations.
In this empirical study we propose an \textit{ad hoc}, lightweight approach that may be adopted to strengthen the existing models.
In future work, several factors that may affect the results, among which are network architecture, regularization, and dropout probability, require further in-depth study.
Another significant improvement would be in a model could estimate the number of samples to acquire on each step.

\section*{Acknowledgements}

The work was supported by the Skoltech NGP Program No.\ 2016-7/NGP (a Skoltech-MIT joint project).


%
%

\renewcommand{\indexname}{Subject Index}

\end{document}